\documentclass[a4paper]{article}

\usepackage{INTERSPEECH2018}

\title{Investigating Speech Features for Continuous Turn-Taking Prediction
Using LSTMs}
\name{Matthew Roddy$^1$, Gabriel Skantze$^2$, Naomi Harte$^1$}
\address{
  $^1$ADAPT Centre, School of Engineering, Trinity College Dublin, Ireland\\
  $^2$Department of Speech Music and Hearing, KTH, Stockholm, Sweden}
\email{roddym@tcd.ie}

\begin{document}

\maketitle
\begin{abstract}
For spoken dialog systems to conduct fluid conversational interactions with users, the systems must be sensitive to turn-taking cues produced by a user. Models should be designed so that effective decisions can be made as to when it is appropriate, or not, for the system to speak. Traditional end-of-turn models, where decisions are made at utterance end-points, are limited in their ability to model fast turn-switches and overlap. A more flexible approach is to model turn-taking in a continuous manner using RNNs, where the system predicts speech probability scores for discrete frames within a future window. The continuous predictions represent generalized turn-taking behaviors observed in the training data and can be applied to make decisions that are not just limited to end-of-turn detection. In this paper, we investigate optimal speech-related feature sets for making predictions at pauses and overlaps in conversation. We find that while traditional acoustic features perform well, part-of-speech features generally perform worse than word features. We show that our current models outperform previously reported baselines.

\end{abstract}

\section{Introduction}

To efficiently communicate with each other in conversations, humans have a tendency towards minimizing the amount of overlap and gap between speaking turns in a conversation \cite{sacks_simplest_1974}. In studies of turn-switch timings, a modal gap of 200ms has been found to exist cross-culturally \cite{stivers_universals_2009}. Considering that this 200ms value is close to the limit of human response time to any stimulus \cite{levinson_timing_2015}, it supports the \emph{projection theory }proposed by Sacks \cite{sacks_simplest_1974}, that efficient communication involves forming predictions with regards to when turn-switches will occur. There have been many studies which have shown how the occurrence and timing of these turn-switches are influenced by an assortment of verbal and non-verbal turn-taking cues such as prosody, semantics, syntax, gesture, and eye-gaze (e.g. \cite{gravano_turn-taking_2011,duncan_signals_1972,kendon_functions_1967}).

When designing spoken dialog systems (SDSs), it is desirable to enable them to pick up on these cues so that they can communicate in a fluid and naturalistic way. Traditionally, SDSs have relied on the use of end-of-turn (EOT) models (e.g. \cite{hariharan_robust_2001,ferrer_is_2002,raux_finite-state_2009}) that wait until a speaker has stopped speaking for a given length of time to form a binary prediction whether the speaker will continue speaking (HOLD) or not (SHIFT). However, this approach is limited in its ability to model naturalistic dialogue for a number of reasons. Firstly, the question of how much context is to be included in the decision-making is open ended. Secondly, in the case of short responses such as backchannels, traditional HOLD/SHIFT decisions may not apply.  Thirdly, the turn-taking behaviors that are modeled by this approach are restricted to HOLD/SHIFT decisions after the pause.  More comprehensive models that can make decisions with a wide variety of turn switch times would allow for more flexible real-world usage. 

An alternative approach proposed in \cite{skantze_towards_2017} is to model turn-taking in a continuous manner using LSTM neural networks. In these models, continuous predictions of a person's speech activity are made at each individual frame step of 50ms. The networks are trained to predict a vector of probability scores for speech activity in each individual frame within a set future window. Rather than designing classifiers to make specific decisions, these continuous models are able to capture general information about turn-taking in the data that they are trained on. They can therefore be applied to a wide variety of turn-taking prediction tasks. They also have the advantage of being able to learn long range dependencies and represent them within the hidden state of the LSTM. These models have been shown to outperform traditional classifiers when applied to HOLD/SHIFT predictions. They have also been shown to outperform humans on the same tasks. However, the questions of which features are most useful for these models and how the features influence network performance on different turn-taking tasks has not yet been thoroughly examined. 

In this paper we significantly extend the original work of Skantze in \cite{skantze_towards_2017}. We perform an investigation of three main speech feature types: acoustic, phonetic, linguistic. We look at the interactions between the different features and propose optimal subsets choices for different tasks. We use sequential forward selection to investigate which of the acoustic features would be most useful for practical implementation. This paper addresses the additional challenge of making turn-taking decisions at points of overlapping speech. Since these models can be applied to a wide variety of turn-taking prediction tasks, this investigation is also of relevance to the study of which modalities are important for different types of turn-taking decisions. Our primary aim in this investigation though is to inform the feature choices in future designs of turn-taking models for SDSs. 

\section{Continuous Turn-taking Prediction}

\subsection{Model Overview}

\begin{figure}[t]
  \centering
  \includegraphics[trim={7cm 0 7cm 0},clip,scale=0.4]{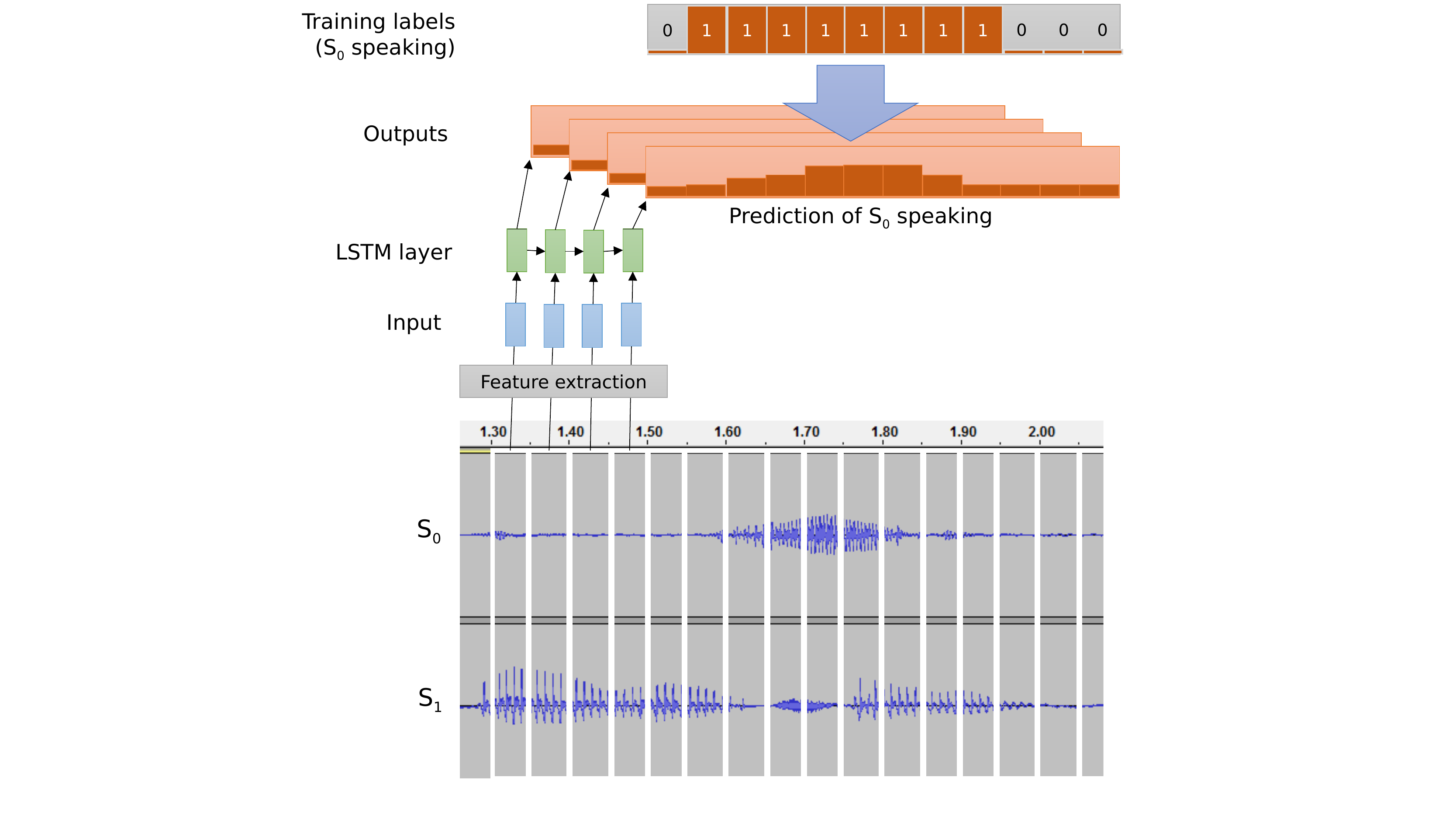}
  \caption{At each 50ms frame, the model makes probabilistic predictions for individual future frames within a window of length N. Figure as in \cite{skantze_towards_2017}}
  \label{fig:predictions}
\end{figure}

Fig. \ref{fig:predictions} shows how LSTM networks are applied to make continuous turn-taking predictions. The main objective is to predict the future speech activity annotations of one of the speakers in a dyadic conversation using input speech features from both speakers ($S_{0}$, $S_{1}$). At each frame ($n$) of size 50ms, speech features are extracted and used to predict the future speech activity of one of the speakers. The future speech activity is a 3 second window comprising of 60 frames of the binary annotations for frames $n+1$ to $n+60$. The output layer of the network uses a sigmoid activation to predict a probability score for the target speaker's speech activity at each future frame. The network uses a single LSTM layer with a variable number of hidden nodes. The features are concatenated into a single feature vector, with the exception of the linguistic features which use an embedding layer that is discussed in section \ref{ling_section}. Each conversation in our data is used twice, with the positions of $S_{0}$ and $S_{1}$ swapped. The networks were trained to minimize binary cross entropy (BCE) loss. Originally in \cite{skantze_towards_2017} mean absolute error (MAE) was proposed as an objective function. We found that using the BCE loss significantly improved the predictive performance of the networks due to BCE directly measuring the difference between class distributions rather than an absolute error distance (this is validated in the results section).  The model was implemented in Python using the PyTorch framework and our code is available online\footnote{www.github.com/mattroddy/lstm\_turn\_taking\_prediction}.

\subsection{Prediction tasks}

We test the performance of our networks using three turn-taking prediction tasks that are pertinent to SDSs. The first two (prediction at pauses, prediction at onset) were proposed in \cite{skantze_towards_2017}. The third, prediction at overlap, is introduced in this paper. 

\paragraph*{Prediction at Pauses}

The prediction at pauses task represents the standard turn-taking decision made at brief pauses in the interaction to predict whether the person holding the floor will continue speaking (HOLD) or the interlocutor will take a turn (SHIFT). To make this decision, we find all points where there is a pause of a minimum set length. We then select all of these instances where only one person continued within a one second window directly after the pause. We average the predicted output probabilities within the window for each of the speakers at the frame directly after the pause. The speaker with the higher score is selected as the predicted next speaker, giving us a binary HOLD/SHIFT classification for which we report F-scores. For all reported F-scores we use 'weighted' F-scores, where the metrics are calculated for both labels and a weighted average is used to take into account imbalance between labels. We test predictions at pauses of both 500ms (PAUSE 500) and 50ms (PAUSE 50). 500ms has been identified in the literature as being a common pause threshold in many commercial speech technologies \cite{heldner_pauses_2010}. 50ms can be considered a threshold that more closely models natural conversational speech since it can potentially generate decisions within the 200ms modal response threshold. The number of samples in the test set for PAUSE 50 is 6533 (4203 HOLDs, 2330 SHIFTs). The number of samples in the test set for PAUSE 500 is 2467 (1496 HOLDs, 971 SHIFTs). The majority baseline F-score (always HOLD) is 0.5037 and 0.4579 for PAUSE 50 and PAUSE 500 respectively. 

\paragraph*{Prediction at Onsets}

Prediction at onsets (ONSET) represents a prediction of the length of an utterance after an initial period of speech. It represents a useful decision for estimating how long the upcoming utterance will be. It categorizes onset predictions into SHORT and LONG, where SHORT utterances can be considered similar to backchannels. For an utterance to be classified as short, 1.5 seconds of silence by a participant has to be followed by a maximum of 1 second of speech, after which the speaker must be silent for at least 5 seconds. For the utterance to be classified as long, 1.5 seconds of silence must be followed by at least 2.5 seconds of speech. The point at which the predictions are made is 500ms from the start of the utterance. The prediction is made by taking the mean of the 60 output nodes from the sigmoid layer and comparing them to a threshold value. During the training stage this threshold is determined by finding the value that best separates the two classes. The number of samples in the test set is 476 (238 SHORTs, 238 LONGs). The majority baseline F-score (always SHORT) is 0.3333.

\paragraph*{Prediction at Overlap }

\begin{figure}[t]
  \centering
  \includegraphics[scale=0.37]{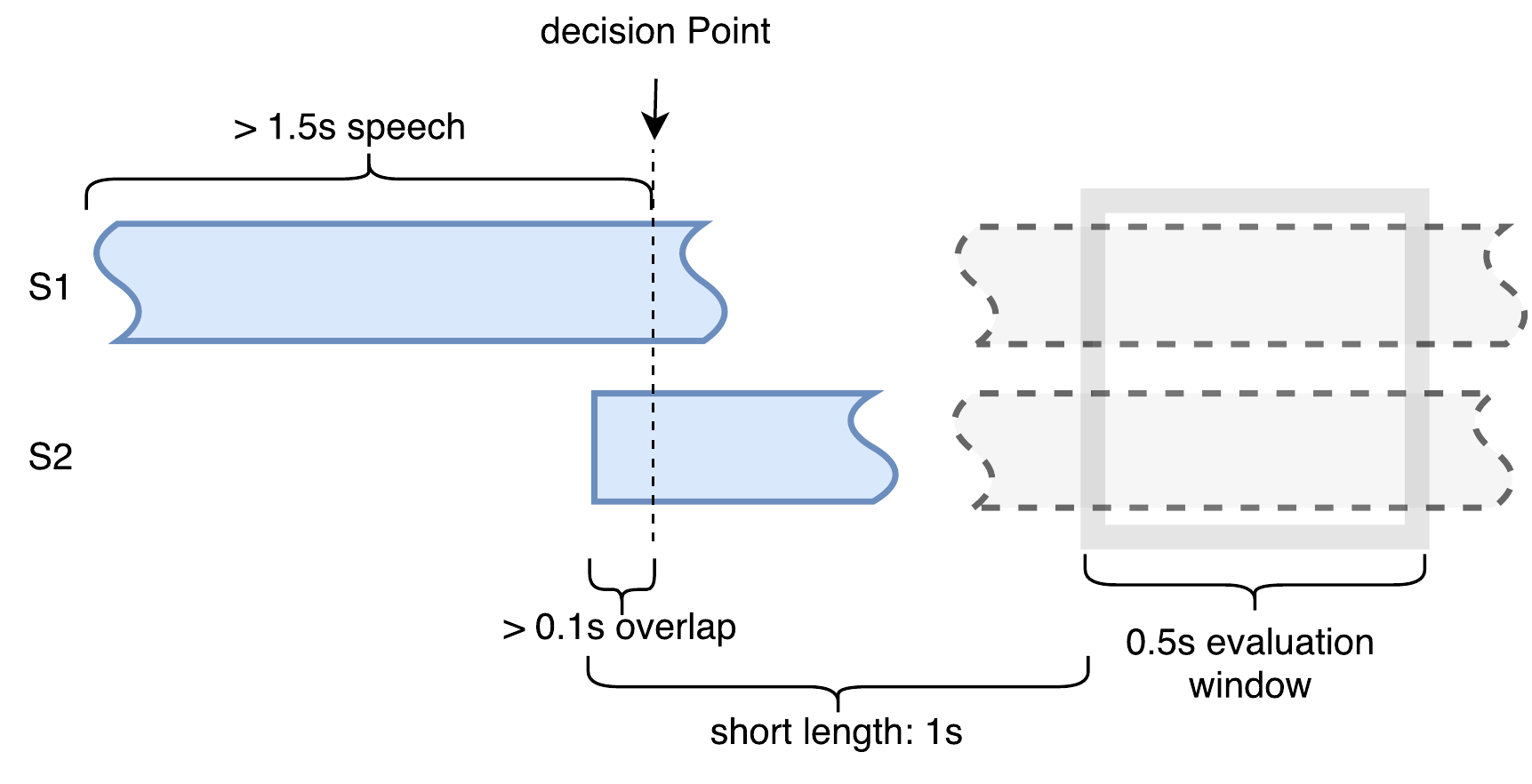}
  \caption{Decisions at overlap.}
  \label{fig:overlap}
\end{figure}

Prediction at overlap (OVERLAP), shown in Fig. \ref{fig:overlap}, is introduced in this paper as a decision to specifically model points where the dialogue system is holding the floor and a user begins speaking while the system is speaking. It makes decisions as to whether it is appropriate for the system to stop speaking or to continue. The decision to continue would be in the case where the overlapping utterance can be considered to be similar to a backchannel. To make this HOLD/SHIFT prediction, decision points are identified were there has been at least 1.5 seconds of speech by a participant that includes a period of 100ms of overlapping speech in the last two frames. From this decision point, 10 frames between 400 and 900 ms in the future are selected as the decision window. If there is only one person speaking we label this decision as either HOLD or SHIFT. To make predictions on this task the means of the output probabilities from the two participants' networks are computed for the 10 frames within the decision window. The two means are then compared to produce a HOLD/SHIFT prediction. The number of samples in the test set is 313 (143 HOLDs, 170 SHIFTs). The majority baseline F-score (always SHIFT) is 0.3823.

\section{Features for Turn-Taking}

\subsection{Acoustic Features}

As acoustic features, we use the low-level descriptors from the eGeMAPs \cite{eyben_geneva_2016} feature set extracted with the OpenSmile toolkit \cite{eyben_opensmile_2010}.  The 21 features in the set include energy (e.g. loudness, shimmer), frequency (e.g. pitch, jitter), and spectral (e.g. spectral flux, MFCCs) features. All the frame-steps and frame-sizes for the sample windows were changed from their default settings to accommodate the 50ms time-step and ensure that there was no overlap of samples between adjacent windows. All other settings were kept the same. The features were then normalized using z-scores on a per-file basis. 

\subsection{Linguistic Features}\label{ling_section}

We investigate two types of linguistic features: Part-of-Speech (POS), and words. POS has been found to be a good predictor of turn-switches in the literature \cite{gravano_turn-taking_2011}. As a comparison, we also use the word transcriptions supplied with the dataset. The comparison is relevant because automatic systems for POS tagging would need the words (from an ASR system) to extract these features. We question whether this extra processing step is indeed necessary, or whether raw word features without the POS tags will suffice. This would be a particular advantage to real-time systems since the added computation of POS tags from the words would be avoided.

For both the words and POS tags, we used the annotations supplied with the data. The number of POS tags was 59 and the number of word tags was 2501. The raw data was represented as an enumerated vocabulary with an added zeroth element representing no change in state. In an effort to simulate the conditions of a real-time system, the linguistic features were not provided to the system until 100ms after the end of the word.

In both cases, the raw features are transformed into separate embeddings of length 64 using added linear network layers that are jointly trained with the rest of the network. The embeddings allow the network to learn representations of the features that are specific for the turn-taking prediction task. Using these fixed length embeddings also makes the comparison between the two features more robust. 

\subsection{Phonetic Features}

For phonetic features we use the bottleneck layer output of a DNN trained to classify senones (tied tri-phone states). In \cite{masumura_online_2017} senone bottleneck features (BNFs) were found to outperform MFCCs in an EOT system based on stacked RNNs. In our implementation, the DNN takes stacked inputs of a central frame with 5 context frames on either side of the central target frame. For each frame 12 MFCCs and their first and second order derivatives are calculated, leading to an input vector of size 396 per target label. To train the DNN we use the senone posterior outputs from a GMM-HMM speech recognition system. The HMM system was trained using a Kaldi recipe on 100 hours of the LibriSpeech corpus \cite{panayotov_librispeech_2015}. We used a DNN with five hidden layers with a bottleneck of size 64 in the fourth layer. The other four hidden layers had 512 hidden nodes and used tanh activation functions. The output layer used a sigmoid activation over the 3448 clustered senone states. Since our ASR system uses frame steps of 10ms and our prediction system uses 50ms frame steps, we use element-wise averaging within each 50ms window. We also delay all of the BNF features by 60ms to compensate for the context that is used in their calculation. 

\subsection{Voice Activity}

The speech transcriptions included with our data were used as a ground-truth for the 60 speech activity predictions. They were also used as a voice activity (VA) feature.

\section{Experimental Setup}

We use the HCRC map task corpus for experiments \cite{anderson_hcrc_1991}, a corpus of 128 dyadic task-based dialogues totaling 18 hours in length. We used 96 conversations as training and 32 conversations for testing. Additionally, we used 32 conversations of the training set as a held-out test set during hyperparameter searches. In selecting the splits between the sets, our main consideration was maintaining speaker independence between the sets. We balanced variables such as gender, and whether the participants could see each other or not. 

We trained models using different combinations of the feature groupings and compared the results. To find optimal hyperparameters for each experimental result, a grid search procedure was used. Networks with hidden node sizes of {[}20,40,60,80,100{]}, learning rates of {[}0.001,0.01{]}, and L2 regularization values of {[}0.001,0.0001{]} were trained and tested on the held-out portions of the training set. Since performance fluctuated between test runs (due to factors such as random weight initialization and randomized batching), once the optimal learning rate and regularization values for a feature set were selected, each hidden node size was trained and tested three times. Once the optimal number of hidden nodes was selected, the network was then trained and tested 10 times on the full train/test split. The averaged F-scores ($f_1$) and losses are shown in Tables \ref{tab:tab_without_va} and \ref{tab:tab_with_va}. The three best results for each metric are shown in bold with the best one in italics. 'Ling' corresponds to the use of both POS and word features. We report the performance of voice activity separately due to its potential masking of the performance of other features. In our discussion we use independent two-tailed t-tests to report on the statistical significance of the difference between the means of metrics.

\begin{figure}[t]
  \centering
  \includegraphics[scale=0.41]{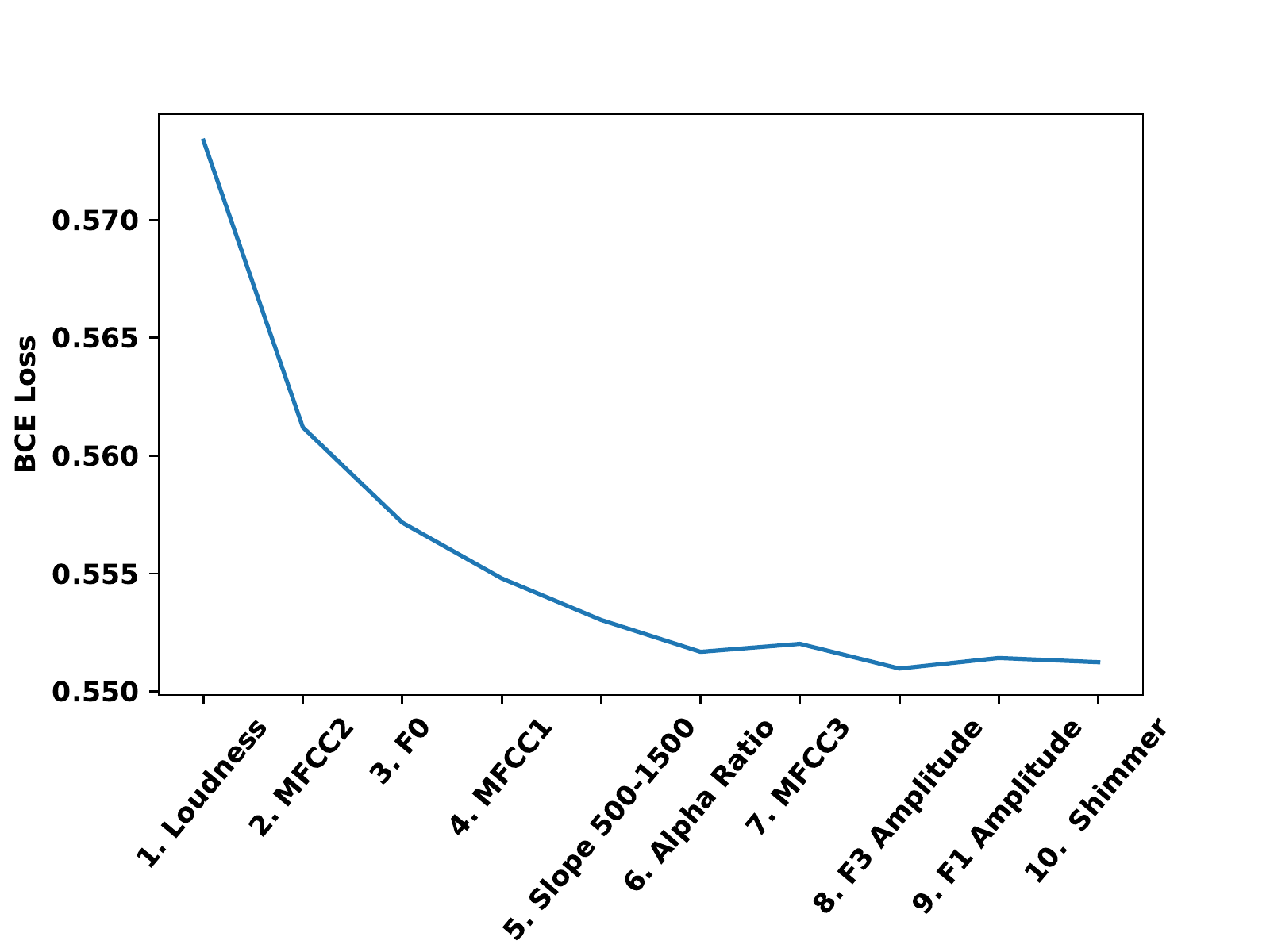}
  \caption{Sequential feature choices with the loss for each consecutively chosen feature.}
  \label{fig:sfs}
\end{figure}

In addition to testing different feature groupings, we run an experiment using the widely used sequential forward selection (SFS) algorithm \cite{kudo_comparison_2000} on the 21 acoustic features to investigate which, and how many, of these features are most useful to turn-taking prediction. During the selection process, at each step of the algorithm we use similar grid-search and testing procedures to those described above. The BCE losses of the first 10 choices are shown in Fig. \ref{fig:sfs}.

\section{Discussion}

\begin{table}[t]
\caption{Performance of feature types (voice activity excluded)}
\label{tab:tab_without_va}
\resizebox{\columnwidth}{!}{
\begin{tabular}{lrrrrr}
\toprule 
{}  & BCE loss  & f1\ 50ms  & f1\ 500ms  & f1\ overlap\  & f1\ onset \tabularnewline
\midrule 
Acous  & \textbf{0.5506}  & 0.7685  & 0.8021  & \textbf{0.6955}  & \textbf{0.782}0 \tabularnewline
Phon  & 0.5645  & 0.7501  & 0.7968  & 0.6670  & 0.7391 \tabularnewline
Pos  & 0.6265  & 0.6454  & 0.6280  & 0.5097  & 0.6015 \tabularnewline
Words  & 0.5842  & 0.7110  & 0.7293  & 0.5965  & 0.7089 \tabularnewline
Ling  & 0.5823  & 0.7165  & 0.7442  & 0.6185  & 0.7125 \tabularnewline
Acous Phon  & 0.5575  & 0.7644  & 0.7924  & 0.6836  & 0.7724 \tabularnewline
Acous Words  & \textbf{\emph{0.5499}}  & \textbf{\emph{0.7811}}  & \textbf{0.8126}  & 0.6876  & 0.7698 \tabularnewline
Acous POS  & 0.5507  & 0.7664  & 0.8031  & \textbf{0.6901}  & \textbf{\emph{0.7841}} \tabularnewline
Acous Ling  & \textbf{0.5502}  & \textbf{0.7793}  & 0.8107  & \textbf{\emph{0.6983}}  & 0.7650 \tabularnewline
Phon Ling  & 0.5589  & 0.7702  & \textbf{0.8128}  & 0.6562  & 0.7505 \tabularnewline
Acous Phon Ling  & 0.5508  & \textbf{0.7788}  & \textbf{\emph{0.8173}}  & 0.6880  & \textbf{0.7768} \tabularnewline
\bottomrule
\end{tabular}

}
\end{table}

\begin{table}[t]
\centering
\caption{Performance of feature types (voice activity included)}
\label{tab:tab_with_va}
\resizebox{\columnwidth}{!}{
\begin{tabular}{lrrrrr}
\toprule 
{}  & BCE loss  & f1\ 50ms  & f1\ 500ms  & f1\ overlap\  & f1\ onset \tabularnewline
\midrule 
VA  & 0.5761  & 0.6800  & 0.7066  & 0.5548  & 0.6281 \tabularnewline
VA Acous  & \textbf{0.5454}  & 0.7926  & 0.8089  & \textbf{\emph{0.7106}}  & \textbf{0.7767} \tabularnewline
VA Phon  & 0.5559  & 0.7765  & 0.8029  & 0.6751  & 0.7426 \tabularnewline
VA POS  & 0.5656  & 0.7374  & 0.7344  & 0.6210  & 0.6939 \tabularnewline
VA Words  & 0.5572  & 0.7708  & 0.7701  & 0.6545  & 0.7216 \tabularnewline
VA Ling  & 0.5573  & 0.7693  & 0.7705  & 0.6506  & 0.7169 \tabularnewline
VA Acous Phon  & 0.5513  & 0.7824  & 0.8034  & \textbf{0.7035}  & 0.7704 \tabularnewline
VA Acous Words  & \textbf{\emph{0.5449}}  & \textbf{\emph{0.7959}}  & \textbf{0.8153}  & 0.6912  & 0.7721 \tabularnewline
VA Acous POS  & \textbf{0.5456}  & 0.7909  & 0.8055  & \textbf{0.7055}  & \textbf{\emph{0.7823}} \tabularnewline
VA Acous Ling  & 0.5461  & \textbf{0.7945}  & 0.8127  & 0.6826  & 0.7674 \tabularnewline
VA Phon Ling  & 0.5505  & 0.7903  & \textbf{\emph{0.8209}}  & 0.6825  & 0.7538 \tabularnewline
VA Acous Phon Ling  & 0.5468  & \textbf{0.7951}  & \textbf{0.8189}  & 0.6943  & \textbf{0.7722} \tabularnewline
\bottomrule
\end{tabular}

}
\end{table}

\subsection{Feature Set Comparison }

A conclusion we can draw from the results in Tables 1 and 2 is that acoustic features are a good first choice for training a system that can perform well on most of our selected turn-taking metrics. All of the best performing feature sets for the studied metrics include acoustic features, with only one exception (the VA Phon Ling combination on PAUSE 500). The added inclusion of linguistic features should be done with consideration for the overall goal of the system and the processing involved in ASR. POS features have been found in previous studies as good turn-taking features \cite{gravano_turn-taking_2011,koiso_analysis_1998}. We find that including word features in addition to acoustic features improves the performance of the networks on the standard turn-taking decisions PAUSE 50 and PAUSE 500 more than POS features ($p<0.05$ in all cases). However, POS features are more useful for the ONSET decisions ($p<0.05$), with the best performance in both tables being achieved with POS tags. So if we intend for the SDS to be able to be able to discern whether an utterance will be short like a backchannel, the results indicate that it would be useful to include POS tags. Concerning phonetic features, the results indicate that their inclusion aids the performance of predictions on the PAUSE 500 metric when used in conjunction with other features. They are less useful at the PAUSE 50 prediction. 

The results imply that good performance on our OVERLAP metric most strongly relies on acoustic features. While the best score on this metric in Table \ref{tab:tab_without_va} is achieved with the combination of acoustic and linguistic features, this mean is not significantly different from the mean of the score for acoustic features alone ($p=0.49$). We therefore suggest that the network is mainly relying on acoustic information in both cases. 

In comparing the results of the two tables, it is perhaps unsurprising that all of the best results on the different metrics are achieved with the inclusion of VA features. The fact that the network can achieve reasonably good performance with just the VA features suggests that it is learning general information about turn distributions within the data. Similar conclusions were drawn in \cite{raux_optimizing_2012} where they found that turn lengths and number of pauses so far in the turn were good predictors of turn-taking behavior.



\subsection{Sequential Forward Selection}

The SFS experiment gives us further insight into which of the acoustic features are the most useful for turn-taking modeling. The plot indicates that after the inclusion of the 8th feature, the loss stops to decrease. This suggests that the most important acoustic features from this set of 21 features are loudness, F0, low order MFCCs, and spectral slope features. F0 and loudness are both traditional features that are established in the literature as indicators of turn-taking behavior (e.g \cite{duncan_signals_1972, gravano_turn-taking_2011}). MFCCs have also been show to be good features for the classification of short listener responses \cite{neiberg_online_2011} while spectral slope features can be considered to contain similar information to these low-order MFCCs. The results indicate that these traditional acoustic features are good choices for continuous turn-taking prediction.

\subsection{Baseline Performance Improvements}

We achieved better performance  than \cite{skantze_towards_2017} on all metrics, apart from the prediction at onset metric. Examination showed that their train/test split was slightly different, and did not use fully speaker-independent sets. When we ran our experiments using their original split, we achieved the best performance on this metric, along with the others but recommend the speaker independent splits as more realistic. To validate our decision to change the loss function from MAE to BCE, we performed a comparison of the two using networks trained on acoustic and word features. All settings were identical except for the loss functions. For PAUSE 50 the mean F-score for the MAE networks was 0.748, while it was 0.7811 for the BCE networks ($p\ll0.001$). In all other F-score metrics we observed similar statistically significant improvements.

\section{Conclusion}

Given that there is disagreement in the literature as to whether linguistic \cite{de_ruiter_projecting_2006} or acoustic \cite{bogels_listeners_2015} information is more important for predicting turn-taking, our findings present evidence that acoustic information is generally more useful as a first choice for continuous turn-taking prediction. Linguistic features can be used to additionally improve performance by using them in conjunction with the acoustic features. The specific choice of which type of linguistic feature to use (e.g. words or POS) should be considered within the context of the demands of the SDS itself, as the utility of features differ based on what type of turn-taking prediction is being made. 

\section{Acknowledgements}
\scriptsize{ The ADAPT Centre for Digital Content Technology is
funded under the SFI Research Centres Programme (Grant
13/RC/2106) and is co-funded under the European Regional
Development Fund}

\bibliographystyle{IEEEtran}

\bibliography{./My_Library.bib}

\end{document}